\documentclass[10pt,twocolumn,letterpaper]{article}

\usepackage{iccv}
\usepackage{times}
\usepackage{epsfig}
\usepackage{graphicx}
\usepackage{graphbox}
\usepackage{amsmath}
\usepackage{amssymb}
\usepackage{booktabs}
\usepackage[table,xcdraw,dvipsnames]{xcolor}
\usepackage{algorithm}
\usepackage[noend]{algpseudocode}

\usepackage[breaklinks=true,bookmarks=false]{hyperref}

\iccvfinalcopy %

\def\iccvPaperID{-} %

\ificcvfinal\fi

\usepackage[capitalize]{cleveref}
\crefname{section}{Sec.}{Secs.}
\Crefname{section}{Section}{Sections}
\Crefname{table}{Table}{Tables}
\crefname{table}{Tab.}{Tabs.}

\def\F1i{$\mathcal{F}1^i$}
\def\owrcnn{OW-RCNN}

\begin{document}

\definecolor{yellow}{HTML}{FFD700}
\definecolor{green}{HTML}{008000}

\title{Addressing the Challenges of Open-World Object Detection}

\author{David Pershouse\\
Queensland University of Technology \\
Brisbane, Australia\\
{\tt\small d3.pershouse@hdr.qut.edu.au}
\and
Feras Dayoub\\
University of Adelaide\\
Adelaide, Australia\\
{\tt\small feras.dayoub@adelaide.edu.au}
\and
Dimity Miller \\
Queensland University of Technology \\
Brisbane, Australia\\
{\tt\small d24.miller@qut.edu.au}
\and
Niko S\"underhauf \\
Queensland University of Technology \\
Brisbane, Australia\\
{\tt\small niko.suenderhauf@qut.edu.au}
}
\maketitle

\ificcvfinal\thispagestyle{empty}\fi

\begin{abstract}

  We address the challenging problem of open world object detection (OWOD), where object detectors must identify objects from known classes while also identifying and continually learning to detect novel objects. Prior work has resulted in detectors that have a relatively low ability to detect novel objects, and a high likelihood of classifying a novel object as one of the known classes. We approach the problem by identifying the three main challenges that OWOD presents and introduce OW-RCNN, an open world object detector that addresses each of these three challenges. OW-RCNN establishes a new state of the art using the open-world evaluation protocol on MS-COCO, showing a drastically increased ability to detect novel objects ($16$-$21\%$ absolute increase in U-Recall), to avoid their misclassification as one of the known classes (up to $52\%$ reduction in A-OSE), and to incrementally learn to detect them while maintaining performance on previously known classes ($1$-$6\%$ absolute increase in mAP). The code will be made publicly available upon acceptance.

\end{abstract}

\section{Introduction}
\label{sec:intro}

\begin{figure}
  \centering

  \includegraphics[width=1\linewidth]{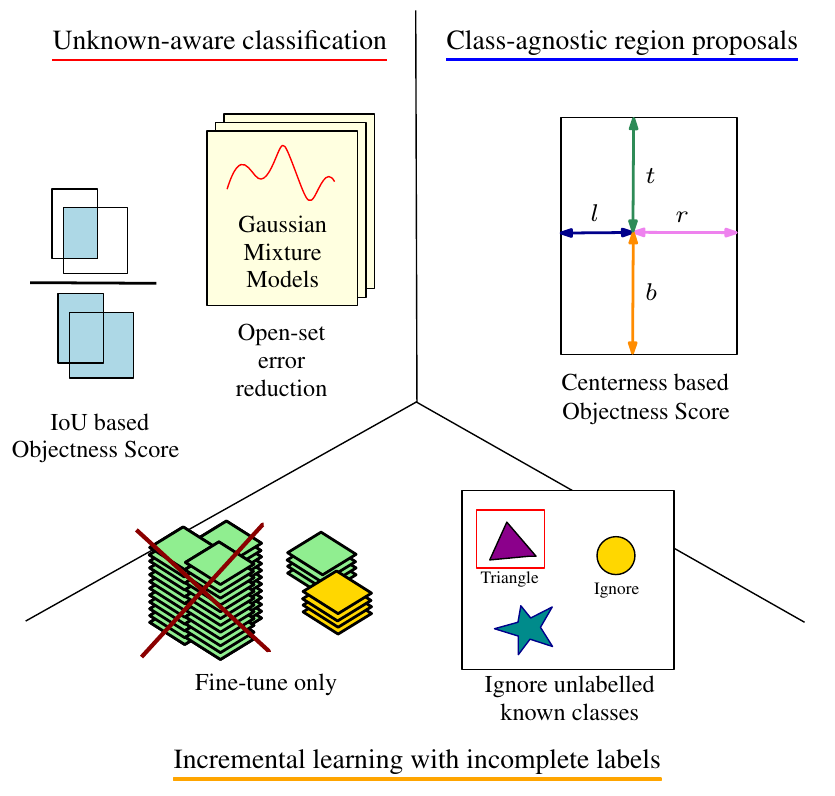}

  \caption{Our solution aims to systematically address each of the core challenges presented by the Open-World Object Detection task. Introducing a centerness based objectness score for generating class agnostic region proposals avoiding the need to learn a classifier for unknown objects, an IoU based objectness score and Gaussian mixture models for unknown aware classification and reducing open-set errors, and a fine-tune-only incremental learning scheme which handles unlabelled known objects in the training data by ignoring them during loss calculation.}
  \label{fig:overview}
\end{figure}

The open-world object detection task (OWOD) was formalised by Joseph \etal~\cite{josephOpenWorldObject2021}, extending the classic object detection task by introducing the requirement to localise objects of unknown classes during inference and incrementally learn from a subset of these objects in future iterations.

Beyond the challenges encountered in standard object detection, we identify three fundamental obstacles that must be tackled by proposed solutions for open-world object detection. These obstacles are: \textbf{1) Class-agnostic region proposals.} The detector must be able to locate and propose regions containing objects while ignoring regions only partially covering an object or containing no objects (background). This task is complicated by the fact that the training data only includes labelled bounding boxes for objects being introduced as known classes. In contrast, unlabelled areas can contain a mix of background, unknown objects, and objects from previously learned classes, further adding to the complexity of the task. \textbf{2) Unknown-aware classification.} Classical object detectors are asked to classify a region as either a known class or background. However, in OWOD, the detector must also account for the possibility of unknown objects being present in a region. This requires two extensions: first, the classifier should have the option to classify a region as unknown, and second, the classifier should protect against open-set errors that are typically triggered by the presence of unknown objects~\cite{dhamijaOverlookedElephantObject2020, millerDropoutSamplingRobust2018}. \textbf{3) Incremental learning with incomplete labels.} In the OWOD setting, an object detector must incrementally learn new object classes without access to the full training data used to learn the previous classes. However, simply fine-tuning the model on the data for the new classes can cause catastrophic forgetting~\cite{frenchCatastrophicForgettingConnectionist1999}, i.e., the model may forget previously learned classes as it adapts to the new data.

Previous OWOD frameworks have been proposed ~\cite{josephOpenWorldObject2021, guptaOWDETROpenworldDetection2022, wu2022uc, maCATLoCalizationIdentificAtion2023, zoharPROBProbabilisticObjectness2022, wuTwobranchObjectnesscentricOpen2022}, each addressing only a subset of the core challenges. When evaluated on MS-COCO for OWOD~\cite{josephOpenWorldObject2021}, these frameworks exhibit low ($22$--$24\%$) detection rates for unknown objects, commonly misclassifying them as one of the known object classes. Furthermore, these frameworks show reduced detection performance after incremental learning of new classes with an absolute drop in mAP between $28\%$ and $31\%$ by the fourth incremental learning task. To address these limitations, this paper presents \owrcnn{}, a novel OWOD framework that systematically addresses each of the identified challenges (see \cref{fig:overview}). OW-RCNN significantly improves the detection performance for unknown objects and reduces the rate of misclassification. Furthermore, our framework improves the retention of detection performance after each incremental learning step compared to existing approaches.

OW-RCNN is built on the Faster R-CNN architecture~\cite{renFasterRCNNRealTime2017}, inspired by its performance under open set conditions ~\cite{dhamijaOverlookedElephantObject2020a}. To achieve \textbf{class-agnostic region proposals} in the presence of unknown objects, OW-RCNN replaces Faster R-CNN's foreground/background classification head with a localization quality regression head. A similar regression head is used to address \textbf{unknown-aware classification} in a novel combination with the existing classification head, providing OW-RCNN with the ability to distinguish between known, unknown, and background regions. To prevent misclassification of unknown objects as a known class, the likelihood of the classifier's logits are modelled for each class by a Gaussian mixture model, with low-likelihood detections treated as unknown.
In order to tackle the challenge of \textbf{incremental learning with incomplete labels}, OW-RCNN adopts a strategy of excluding from training any unlabelled image region that the model classifies as belonging to a previously known class. Furthermore, we reduce catastrophic forgetting by utilising a retraining process based on a pre-trained baseline and a small set of exemplars.

\textbf{Our contribution:} We present \owrcnn{}, a strong open world object detection baseline that addresses the three identified OWOD challenges. We demonstrate the effectiveness of \owrcnn{} by comparing it with previous work using the open-world evaluation protocol~\cite{josephOpenWorldObject2021} on MS-COCO~\cite{linMicrosoftCOCOCommon2014}, revealing an absolute increase of up to 21\% in the unknown recall metric (U-Recall), up to 6\% in the mean average precision metric (mAP), and up to a 52\% reduction in absolute open-set errors (A-OSE).

\section{Related work}
Open-world object detection (OWOD) was formalised recently by Joseph \etal~\cite{josephOpenWorldObject2021}, who also introduced ORE as the first solution to this task. Gupta \etal built upon this to propose an Open World Detection Transformer (OW-DETR) ~\cite{guptaOWDETROpenworldDetection2022}. Recent works extend the trend of using a transformer with Ma \etal~\cite{maCATLoCalizationIdentificAtion2023} introducing the Localization and Identification Cascade Detection Transformer (CAT) and Zohar \etal~\cite{zoharPROBProbabilisticObjectness2022} proposing a probabilistic approach for open-world object detection (PROB). Wu \etal's~\cite{wuTwobranchObjectnesscentricOpen2022}'s two-branch objectness-centric detector (2B-OCD) finds continued utility with a Faster R-CNN~\cite{renClassincrementalObjectDetection} based detector, as does UC-OWOD proposed by
Wu \etal~\cite{wu2022uc} in an extension of the OWOD task, where detected unknown objects should be clustered into their respective unknown classes. Each framework offers different approaches and addresses different subsets of our identified challenges, as we will discuss below.

\textbf{Class-agnostic object proposals} allow an OWOD technique to distinguish between objects of interest and background regions, isolating the regions on which to perform \textbf{unknown-aware classification}. ORE~\cite{josephOpenWorldObject2021}, OW-DETR~\cite{guptaOWDETROpenworldDetection2022}, UC-OWOD~\cite{wu2022uc}, and CAT ~\cite{maCATLoCalizationIdentificAtion2023} approach this by using a foreground-vs-background classifier.
These methods identify the top-k objectness regions which are unlabelled in a training image and pseudo-label them as unknown objects. The pseudo-labels are used as supervision to train a classifier with additional `unknown' outputs. As recently shown by Kim \etal ~\cite{kimLearningOpenWorldObject2022}, this classifier-based approach to class-agnostic detection shows poor generalisation, overfitting on the training categories and not extending to unknown object classes. 2B-OCD~\cite{wuTwobranchObjectnesscentricOpen2022} and PROB~\cite{zoharPROBProbabilisticObjectness2022} avoid the use of pseudo-labels and instead rely on a regression-based class-agnostic objectness score that is used during unknown-aware classification to discriminate between unknown objects or background. Although recent work on class-agnostic detection has taken advantage of instance segmentation masks during training ~\cite{saito2022learning, wang2022open}, segmentation masks are not provided in the OWOD task and are labour intensive to generate. To address class-agnostic proposals in OW-RCNN, we adopt localisation quality regression heads ~\cite{kimLearningOpenWorldObject2022, tianFCOSFullyConvolutional2019} that circumvent the drawbacks associated with the classifier-based approach used in many prior OWOD frameworks.

As established in prior work~\cite{dhamijaOverlookedElephantObject2020, millerDropoutSamplingRobust2018}, unknown objects typically trigger open-set errors, where the object is misclassified as a known class. Of the existing OWOD techniques, only ORE explicitly addresses \textbf{open-set error correction for unknown-aware classification}, where it uses a combination of energy functions and distributions to model the logits generated by either unknown or known objects~\cite{josephOpenWorldObject2021}. However, the distributions need to be trained on labelled examples of unknown objects, which conflicts with the underlying assumptions of the OWOD problem. In the related field of open-set object detection, the handling of open-set errors has previously been explored~\cite{millerDropoutSamplingRobust2018, millerEvaluatingMergingStrategies2019a, dhamijaOverlookedElephantObject2020}, with Miller \etal~\cite{millerUncertaintyIdentifyingOpenSet2022} showing the utility of Gaussian mixture models for detecting open set errors. Uniquely among OWOD frameworks, \owrcnn{} builds upon this related work to specifically address open-set error.

To avoid catastrophic forgetting that can occur during \textbf{incremental learning}, the previous OWOD frameworks ~\cite{josephOpenWorldObject2021, guptaOWDETROpenworldDetection2022, wu2022uc, maCATLoCalizationIdentificAtion2023, zoharPROBProbabilisticObjectness2022, wuTwobranchObjectnesscentricOpen2022} use an exemplar replay strategy where the detection network is fine-tuned on a small, class-balanced subset of images from the current and prior tasks after training on the current task's data. The PROB framework is unique in that it prefers to select images with very high or very low objectness in its subset. Continual learning is an open problem for deep neural networks, and strategies that involve regulating weight updates of a network have been proposed in ~\cite{kirkpatrickOvercomingCatastrophicForgetting2017,maltoniContinuousLearningSingleIncrementalTask2019}, while ~\cite{lampertIncrementalClassifierRepresentation, lopez-pazGradientEpisodicMemory2017} suggest augmenting these regularisation approaches by using small amounts of replay memory. Shmelkov \etal~\cite{shmelkovIncrementalLearningObject2017} reduces catastrophic forgetting of an object detector while incrementally learning new classes by using a knowledge distillation~\cite{hintonDistillingKnowledgeNeural2015a} approach. However, Prabhu \etal~\cite{prabhuGDumbSimpleApproach2020} have subsequently shown that memory replay alone may achieve comparable results to more complicated methods.

None of the examined frameworks address the possible presence of incomplete data labelling in the OWOD task, where unlabelled objects of known classes may exist in the training data. In a related field, Liu \etal~\cite{liuIncDetDefenseElastic2020} address the problem of \textbf{incomplete labels} in continual learning by using the the prior task's detection model to pseudo-label instances of previously known classes that are detected in the current iteration's training data. In our \owrcnn{} framework, we address incomplete labels by using the existing model to detect instances of previously known classes and excluding them from the training data, avoiding the usage of potentially inaccurate pseudo-labels.

Having framed OW-RCNN in the context of previous work we proceed to introduce the details of our framework.

\begin{figure*}
  \centering

  \includegraphics[width=1\linewidth]{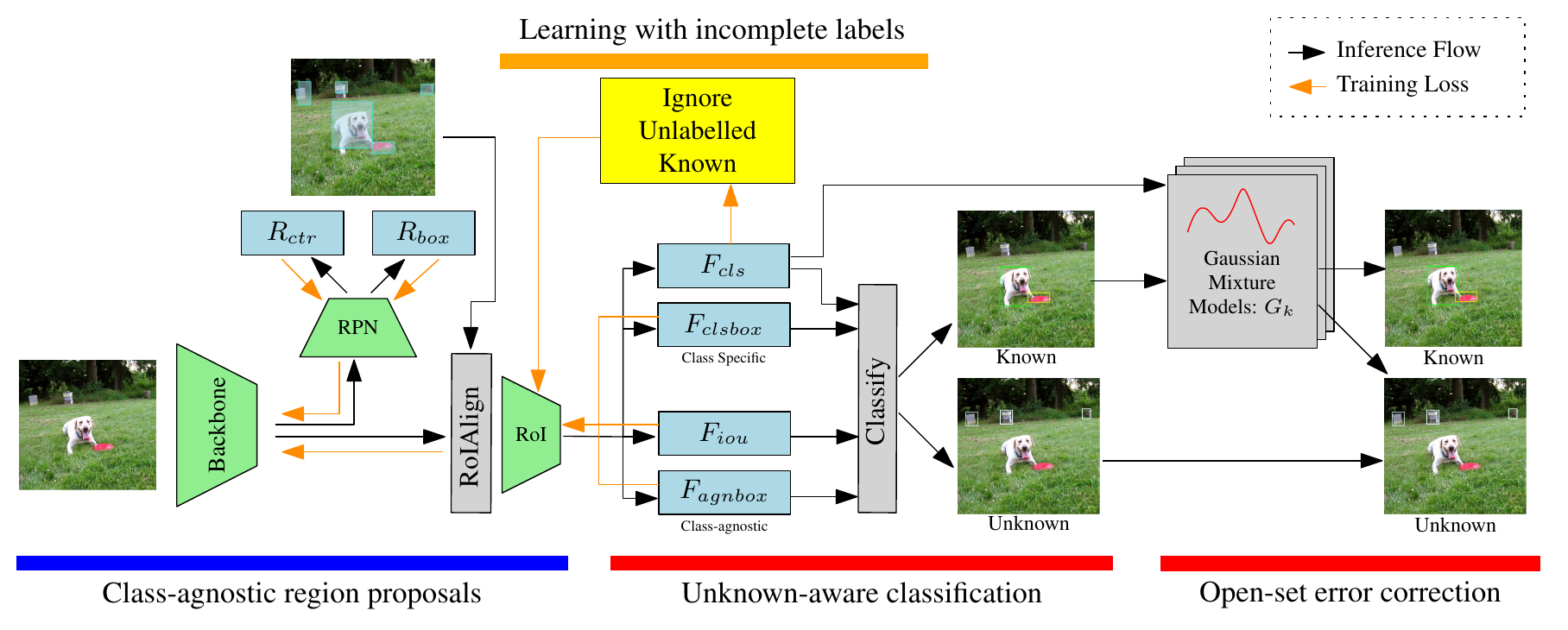}

  \caption{\textbf{The \owrcnn{} architecture.} Our architecture extends Faster R-CNN~\cite{renFasterRCNNRealTime2017} to address the challenges presented by open-world object detection. Its region proposal network's classification-based objectness head is replaced with a localisation-quality based centerness head $R_{ctr}$, predicting the centerness of the proposal boxes and using this to provide class-agnostic region proposals. To the region of interest (ROI) heads, we add a class-agnostic box regression head $F_{agnbox}$ and a head to predict its IoU with the ground truth $F_{iou}$, using both heads to assist with unknown-aware classification. The likelihoods of detected known classes are validated against a Gaussian mixture model to reduce open-set errors. During training confident class predictions for previously known classes that are not labelled in the training data, are ignored when computing loss for the classification head. The network is trained end-to-end using cross-entropy loss for the classification head $F_{cls}$, and $\ell_1$ losses for the other five heads. }
  \label{fig:architecture}
\end{figure*}
\section{Introducing \owrcnn{}}
\subsection{Preliminaries}
We can formalise the open-world detection problem as an object detection problem occurring over many sequential tasks $t$. For each task, there is a set of object classes $C^t \subset \mathcal{C}$ where $\mathcal{C}$ is the set of all possible object classes and a set of training data $\mathcal{D}^t = \{ \mathcal{X}^t, \mathcal{Y}^t \}$ where $\mathcal{X}^t$ is a set of $n$ training images $\{ X_i, X_2, \ldots,  X_n \}$ and $\mathcal{Y}^t$ is the set of annotations $\{ y_1, y_2, \ldots, y_n \}$ corresponding to the objects in each image. Each annotation $y_i$ encodes the location ($x_i, y_i, w_i, h_i$) of an object in the image and its object class $c_i$ i.e., $y_i = \{ c_i, x_i, y_i, w_i, h_i \}$ where annotations exist only for object classes that are part of the current task, that is, $c_i \in C^t$. A model $\mathcal{M}^t$ is trained using the training data $\mathcal{D}^t$, the previous model $\mathcal{M}^{t-1}$ and optionally a small subset of the previous training data $\mathcal{R}^t = \{ R^1 \cup  R^2 \cup \ldots \cup R^{t-1} \}$ where $R^n \subset \mathcal{D}^n$ and $\vert R^n \vert \ll \vert \mathcal{D}^n \vert$.

For a given image $I$, the detector must detect the regions $\{ c_i, x_i, y_i, w_i, h_i \}$ corresponding to objects of known classes $\mathcal{K}^t = C^1 \cup C^2 \cup \ldots \cup C^t$, and a set of regions $\{ x_i, y_i, w_i, h_i \}$ of objects of unknown class $\mathcal{U}^t = \mathcal{C} - \mathcal{K}^t$. The set of unknown regions is forwarded to an oracle, which may annotate a subset of them with classes from $C^{t+1}$, adding the image and the resulting annotations to the training data for the next task $\mathcal{D}^{t+1}$. The new training data is used to produce a new model $\mathcal{M}^{t+1}$ that can detect old and new classes $\mathcal{K}^{t+1} = \mathcal{K}^t + C^{t+1}$ and the remaining unknowns $\mathcal{U}^{t+1} = \mathcal{C} - \mathcal{K}^{t+1}$. This sequence of learning to detect new object classes from previous observations repeats indefinitely.

\subsection{Our architecture}
\label{sec:ourarchitecture}
Our architecture (\cref{fig:architecture}) is an adaptation of Faster R-CNN, with changes and additions designed to address each of the three identified challenges introduced by open-world object detection.

To produce class-agnostic region proposals, an image proceeds through a backbone and the resulting features are processed by the Region Proposal Network (RPN), which consists of a box regression head $R_{box}$ and a localisation quality regression head $R_{ctr}$ that predict the centerness of the bounding box predicted by the box head. The highest scoring proposals based on the centerness are used as regions of interest and will contain a mixture of known and unknown objects.

To perform an unknown-aware classification of the proposed regions into a known class, an unknown class, or background, the features of the class-agnostic proposals are extracted from the feature maps and resized using RoIAlign~\cite{renFasterRCNNRealTime2017}. They are sent through the two fully connected layers of the region of interest (RoI) network before branching into four separate heads. The class-specific classification head $F_{cls}$ and the class-specific bounding box regression head $F_{clsbox}$ are used in combination with the class-agnostic bounding box $F_{agnbox}$ and the IoU-based localisation quality $F_{iou}$ heads, providing input for our unknown classification algorithm (\cref{algo:getclassprobs}).
Our architecture employs Gaussian Mixture Models ($G_k$) to estimate the likelihood of the classification head's logits belonging to each known class ($\mathcal{K}^t$), and this likelihood is used to reduce open-set errors in the classification process (\cref{algo:handleopenset}).

To meet the incremental learning challenge of the open-world object detection problem, and avoid catastrophic forgetting, we train a new model for each task from starting from a pretrained base model using a set of exemplars for both the previous and current task's classes.  During training, we use the classification head $F_{cls}$ and the localisation quality heads $F_{iou}$, $R_{ctr}$ to detect and exclude possible unlabelled instances of previous classes from the loss calculations.

We now detail each of our architecture's components and training procedure and how they meet the challenges introduced by open-world object detection.

\subsection{Class-agnostic region proposals}

The first challenge of open-world object detection is learning to locate and propose regions that contain objects while ignoring regions that contain background. Our architecture is built on Faster R-CNN, which provides us with a strong baseline object detection performance while maintaining a high inference speed. Due to the potentially large variability of the scale of unknown objects, we include a Feature Pyramid Network~\cite{linFeaturePyramidNetworks2017} (FPN). Faster R-CNN uses a dedicated Region Proposal Network (RPN) to achieve its high detection performance. However, when the network is trained only on images annotated with known classes, it cannot detect unknown objects. To address this challenge and to make the RPN produce unknown-aware region proposals, our box regression head $R_{box}$ is trained to predict the distances from the anchor centre to the left, top, right, and bottom edges $(l,t,r,b)$ of the ground truth bounding box. We avoid having to classify between unknown and background by using a regression-based localisation quality head $R_{ctr}$~\cite{kimLearningOpenWorldObject2022}, which is trained to predict the centerness~\cite{tianFCOSFullyConvolutional2019} (\cref{eq:centerness}) of the box head's ($R_{box}$) output.

\begin{equation}
  centerness =  \sqrt{\dfrac{\min \left( l,r \right) }{\max \left( l,r \right)} \times \dfrac{\min \left( t,b \right) }{\max \left( t,b \right)} }
  \label{eq:centerness}
\end{equation}

During training, the RPN centerness head $R_{ctr}$ is trained on a random selection of anchors with centres located within the ground truth bounding boxes. By learning to predict the quality of our bounding box prediction, we allow the network to learn features from across all classes in the training data, learning a generic and class-agnostic objectness score. By only using labelled regions during training we also avoid potentially inaccurate pseudo-labels used by previous frameworks.

At inference time, we take the top proposals ranked by centerness, extract their features from the feature maps, and pass the resulting proposals to the Region of Interest (RoI) network to undergo unknown-aware classification.

\subsection{Unknown-aware classification}
\label{sec:unkclass}
The second challenge of open-world object detection is classifying a proposal as a known object class, an unknown object, or as background, while avoiding open-set errors traditionally caused by the presence of unknown objects.

Unlike most prior work which adds an output to the classifier to handle the unknown class. Our solution treats the role of the classification head in \owrcnn{} as determining whether the region contains a known object or $other$ (a combination of $background$ and $unknown$).

To help determine whether a region detected as $other$ is an $unknown$ object or $background$, we employ a class-agnostic bounding box regression head $F_{agnbox}$ and a localisation-based regression head $F_{iou}$ which is trained to predict the Intersection over Union (IoU) of the bounding box predicted by $F_{agnbox}$ and an object.

We calculate an objectness score $s_{obj}$ from the localisation quality heads, $F_{iou}$ and $R_{ctr}$, using \cref{eq:objectness} ~\cite{kimLearningOpenWorldObject2022}.

\begin{equation}
  s_{obj} = \sqrt{\sigma\left( R_{ctr} \right)\times \sigma\left( F_{iou} \right) }
  \label{eq:objectness}
\end{equation}

We combine the objectness score with the output of the other heads using \cref{algo:getclassprobs} to produce a set of per-class scores and predicted bounding boxes for each of the known classes $\mathcal{K}^t$ together with the $unknown$ and $background$ classes.

\begin{algorithm}[t]\small
  \caption{\textsc{Calculate Class Scores And Boxes}}
  \label{algo:getclassprobs}
  \begin{algorithmic}[1]
    \Require{Classification logits for known and background classes: $F_{cls}$; Class-specific box predictions: $F_{clsbox}$; Class-agnostic box prediction: $F_{agnbox}$; Objectness score: $s_{obj}$; Objectness threshold: $\theta_{obj}$; Low-confidence threshold: $\theta_{conf}$; Known classes: $\mathcal{K}^t$
    }

    \State $S_{cls} \leftarrow softmax\left( F_{cls} \right)$
    \State $S_{cls}[unknown] \leftarrow 0$ \Comment{Create a score for unknown set to 0}
    \State $B_{cls} \leftarrow boxes\left( F_{clsbox} \right)$
    \State $B_{cls}[unknown] \leftarrow boxes\left( F_{agnbox}\right)$
    \If{$S_{cls}[k] < \theta_{conf}$ $\forall$ $k \in \mathcal{K}^t$}
    \If{$s_{obj} > \theta_{obj}$}
    \State $S_{cls}[unknown] \leftarrow s_{obj}$
    \State $S_{cls}[background] \leftarrow 0$ \Comment{Zero the score given to the background class}
    \EndIf
    \EndIf

  \end{algorithmic}
\end{algorithm}

The bounding box for the $unknown$ class is set to the class-agnostic bounding box of $F_{agnbox}$, while its class score is initially set to zero. If the scores for all known classes are of low confidence $\theta_{conf}$, while the objectness score $s_{obj}$ is above a threshold $\theta_{obj}$, then the unknown score is set to the objectness score $s_{obj}$ and the background score resets to zero.

An advantage of using the class-agnostic heads to detect unknown objects is that features used to train the heads come from all classes in the training data, helping the network to learn a general representation of an object and improving our ability to detect unknown objects not encountered in the training data.

Although the objectness score can help us classify an object into a known or unknown class, previously unseen objects may still be incorrectly confidently classified as one of the known classes. To address the second part of this challenge and reduce open-set errors, we utilise Gaussian mixture models to determine the likelihood of the detection network's classification output. A Gaussian mixture model $G_k$ is created and stored per known class. They are trained on the logits produced by the classification head of the detection network $F_{cls}$ for each instance of the class in the training data while using its ground truth bounding box as the region proposal. The lowest likelihood produced by each mixture model for its training data becomes the class specific likelihood threshold $\theta_{like}$. The models are used during inference to detect when the detection network has made an overconfident prediction (\cref{algo:handleopenset}).

\begin{algorithm}\small
  \caption{\textsc{Handle Overconfident Prediction}}
  \label{algo:handleopenset}
  \begin{algorithmic}[1]
    \Require{Classification logits: $F_{cls}$;  Objectness score: $s_{obj}$; class score threshold: $\theta_{cls}$; The predicted class: $k$, its score $s_{cls}$, its Gaussian mixture model $G_k$, and its likelihood threshold: $\theta_{like}$.
    }
    \If{$k \in \mathcal{K}^t$ and $s_{cls} < \theta_{cls}$}
    \State $s_{like} \leftarrow G_k\left( F_{cls} \right)$

    \If{$s_{like} < \theta_{like}$}
    \State $k \leftarrow unknown$
    \State $s_{cls} \leftarrow s_{obj}$
    \EndIf
    \EndIf

  \end{algorithmic}
\end{algorithm}

For a region to change from $known$ to $unknown$, it must first have been classified as a known class with a class score $s_{cls}$ below a threshold $\theta_{cls}$. By applying a threshold to the classification score, we aim to prevent inaccuracies in the mixture model from converting high-scoring regions into $unknown$. The logits $F_{cls}$ of the region have their likelihood $s_{like}$ calculated by the mixture model $G_k$. Regions with likelihoods $s_{like}$ lower than a class-specific threshold $\theta_{like}$, are assumed to be an open-set error; consequently, the detected class is changed to $unknown$ and the class score $s_{cls}$ becomes the objectness score $s_{obj}$. The predicted bounding box remains the one predicted by $F_{clsbox}$.

By using object scores to separate unknown objects from the background and Gaussian mixture models to reduce open-set errors, \owrcnn{} addresses the second challenge introduced by the open-world detection problem.

\subsection{Incremental learning with incomplete labels}

The third challenge that our framework needs to address is incremental learning of new classes. Existing open-world detection frameworks use an exemplar replay approach to achieve incremental learning of new classes, where the model created during the previous task, $\mathcal{M}^{t-1}$, is first trained on the new training data $D^{t}$ and then fine-tuned on a balanced set of exemplars $\mathcal{R}^t$. Although some frameworks show strong incremental learning performance when applying this method to the single task benchmark proposed by Shmelkov \etal~\cite{shmelkovIncrementalLearningObject2017}, under the open-world evaluation protocol, the results show a strong preference for performance on the previous task's classes, with current task class performance as low as half of the previously known classes. We present an alternative method which starts with a pre-trained base model $\mathcal{M}^0$ for each task and fine-tunes with only the set of exemplars $\mathcal{R}^t$ which has been modestly increased in size. This results in a reduced performance drop over higher task counts, with the added benefit of reducing training time and compute costs. The set of exemplars $\mathcal{R}^t$ is built using the same class-balanced sampling algorithm as described by Joseph \etal~\cite{josephOpenWorldObject2021} (see \cref{algo:subsample}) and involves sampling images until each class has a minimum number of instances.

\begin{algorithm}[t]\small
  \caption{\textsc{Build Exemplar Set $R^t$ for Task $t$}}
  \label{algo:subsample}
  \begin{algorithmic}[1]
    \Require {Data for current task: $\mathcal{D}^t$; Valid classes for the current task $C^t$, Minimum number of instances to collect per class $n$ }
    \Ensure  {Sampled images $S$}
    \\ \textit{Initialise a first-in-first-out queue for each class}
    \For {$class \in C^t$}
    \State $M[class] \leftarrow queue(max\_size=n)$
    \EndFor

    \textit{add each image instance to the relevant queue}
    \For {$image \in \mathcal{D}^t$}
    \For {$instance \in image$}
    \State {$M[class] \leftarrow image$}
    \EndFor
    \If {length($M[class]$) $ \geq x\ \forall\ class \in C^t$)}
    \State $R^t \leftarrow unique(M[class]\ \forall\ class \in C^t)$
    \State \Return  $R^t$
    \EndIf
    \EndFor

  \end{algorithmic}
\end{algorithm}

By only keeping a small exemplar set, $R^n$, we satisfy the incremental learning requirement that $\vert R^n \vert \ll \vert D^n \vert$. Our framework trains using the entire set of data $D^1$ during the first task while using only the exemplar set $\mathcal{R}^n$ for subsequent tasks. Although there remains a significant performance gap compared to a model trained on all data, in experiments, we find that it provides balanced performance between current and previous task classes, reduces training time, and improves performance compared to the method used by prior OWOD frameworks.

\subsection{Training}

\owrcnn{}'s detection network is trained by reducing the combined loss of each of the six network heads:

\begin{equation}
  \begin{split}
    \mathcal{L} = & \mathcal{L}_{ctr}+\mathcal{L}_{box} + \\
    &  3\mathcal{L}_{cls}+\mathcal{L}_{clsbox}+\mathcal{L}_{agnbox}+\mathcal{L}_{iou}
  \end{split}
\end{equation}

With all losses calculated using $\ell_1$ loss, except $\mathcal{L}_{cls}$, which uses cross-entropy loss, we found that applying a weighting to the classification loss helps to retain classification importance in the presence of the newly introduced loss terms. After training the detection network for a task, the Gaussian mixture models and likelihood thresholds for each known class are recreated based on the training data ($D^1$ for the first task, $R^t$ for the remainder). The thresholds $\theta_{like}$ are calculated by taking the minimum value of the likelihoods generated by the training data for each class.

An additional challenge introduced by the open-world object detection task is the incomplete labelling of known classes in the training data. The problem formulation stipulates that when training on the task $T^n$, the only labels present in the training set are for the classes introduced as part of the task, $C^n$. For any image in the task $T^n$ there may be objects present from $\mathcal{K}^{n-1}$ that are not labelled. Without explicit handling, a detector's performance on these unlabelled objects may be reduced as it learns to classify them as either background or unknown. To avoid this mixed training signal, we exclude from the loss calculations for the classification head $F_{cls}$ any region that has a confident prediction $s_{cls} > \theta_{cls}$ for a previously known class, a high objectness score $s_{obj} > \theta_{obj}$ and does not overlap with any ground-truth labels.

After describing our framework and how it addresses each of the challenges introduced by the open-world detection task, we evaluate its performance against the other OWOD implementations.

\begin{table*}[pthp]\setlength{\tabcolsep}{2pt}
  \caption{\textbf{State-of-the-art for the Open World Evaluation Protocol on MS-COCO.} We compare our framework with other proposed implementations. Following ~\cite{guptaOWDETROpenworldDetection2022}, we report performance for ORE~\cite{josephOpenWorldObject2021} without the energy-based unknown identifiers. The mAP figures are presented for previously known classes, those that were learnt as part of the current task, and for all known classes. The unknown class recall (U-Recall) represents the portion of unknown objects retrieved by the model. \owrcnn{} shows significant gains in all metrics, for all tasks, highlighting its improved ability to address the challenges of open-world object detection. }\vspace{0.2cm}
  \resizebox{\textwidth}{!} {
    \begin{tabular}{@{}lcccccccccccccccc@{}}
      \toprule

      Task IDs ($\rightarrow$)                                                      &
      \multicolumn{2}{c}{Task 1}                                                    &
      \multicolumn{4}{c}{Task 2}                                                    &
      \multicolumn{4}{c}{Task 3}                                                    &
      \multicolumn{3}{c}{Task 4}                                                                                                                                                                                                                        \\

      \cmidrule(lr){2-3}\cmidrule(lr){4-7}\cmidrule(lr){8-11}\cmidrule(lr){12-14}   &
      U-Recall                                                                      & \multicolumn{1}{c}{mAP ($\uparrow$)}                                                 &
      U-Recall                                                                      & \multicolumn{3}{c}{mAP ($\uparrow$)}                                                 &
      U-Recall                                                                      & \multicolumn{3}{c}{mAP ($\uparrow$)}                                                 &
      \multicolumn{3}{c}{mAP ($\uparrow$)}                                                                                                                                                                                                              \\

      \cmidrule(lr){3-3} \cmidrule(lr){5-7} \cmidrule(lr){9-11}\cmidrule(lr){12-14} &
      ($\uparrow$)                                                                  & \begin{tabular}[c]{@{}c}Current \\ known\end{tabular} &
      ($\uparrow$)                                                                  & \begin{tabular}[c]{@{}c@{}}Previously\\  known\end{tabular}                          & \begin{tabular}[c]{@{}c@{}}Current \\ known\end{tabular} & Both          &
      ($\uparrow$)                                                                  & \begin{tabular}[c]{@{}c@{}}Previously \\ known\end{tabular}                          & \begin{tabular}[c]{@{}c@{}}Current \\ known\end{tabular} & Both          &
      \begin{tabular}[c]{@{}c@{}}Previously \\ known\end{tabular}                   & \begin{tabular}[c]{@{}c@{}}Current \\ known\end{tabular}                             & Both                                                                       \\

      \midrule

      Faster R-CNN\scriptsize{(+threshold)}                                         &
      17.9                                                                          & 61.5                                                                                 &
      13.9                                                                          & 44.7                                                                                 & \textbf{45.0}                                            & 44.8          &
      19.2                                                                          & 33.3                                                                                 & \textbf{36.3}                                            & 34.3          &
      29.3                                                                          & 27.8                                                                                 & 28.9                                                                       \\

      \midrule

      ORE - EBUI~\cite{josephOpenWorldObject2021}                                   &
      4.9                                                                           & 56.0                                                                                 &
      2.9                                                                           & 52.7                                                                                 & 26.0                                                     & 39.4          &
      3.9                                                                           & 38.2                                                                                 & 12.7                                                     & 29.7          &
      29.6                                                                          & 12.4                                                                                 & 25.3                                                                       \\

      OW-DETR~\cite{guptaOWDETROpenworldDetection2022}                              &
      7.5                                                                           & 59.2                                                                                 &
      6.2                                                                           & 53.6                                                                                 & 33.5                                                     & 42.9          &
      5.7                                                                           & 38.3                                                                                 & 15.8                                                     & 30.8          &
      31.4                                                                          & 17.1                                                                                 & 27.8                                                                       \\

      UC-OWOD~\cite{wu2022uc}                                                       &
      -                                                                             & 50.7                                                                                 &
      -                                                                             & 33.1                                                                                 & 30.5                                                     & 31.8          &
      -                                                                             & 28.8                                                                                 & 16.3                                                     & 24.6          &
      25.6                                                                          & 15.9                                                                                 & 23.1                                                                       \\

      2B-OCD~\cite{wuTwobranchObjectnesscentricOpen2022}                            &
      12.1                                                                          & 56.4                                                                                 &
      9.44                                                                          & 51.6                                                                                 & 25.3                                                     & 38.5          &
      11.7                                                                          & 37.2                                                                                 & 13.2                                                     & 29.2          &
      30.1                                                                          & 13.3                                                                                 & 25.8                                                                       \\

      CAT~\cite{maCATLoCalizationIdentificAtion2023}                                &
      21.8                                                                          & 59.8                                                                                 &
      19.2                                                                          & 54.6                                                                                 & 32.6                                                     & 43.6          &
      24.4                                                                          & 42.3                                                                                 & 18.9                                                     & 34.5          &
      34.4                                                                          & 16.6                                                                                 & 29.9                                                                       \\

      PROB~\cite{zoharPROBProbabilisticObjectness2022}                              &
      19.4                                                                          & 59.5                                                                                 &
      17.4                                                                          & \textbf{55.7}                                                                        & 32.2                                                     & 44.0          &
      19.6                                                                          & 43.0                                                                                 & 22.2                                                     & 36.0          &
      35.7                                                                          & 18.9                                                                                 & 31.5                                                                       \\

      \midrule

      \textbf{\owrcnn{} (Ours)}                                                     &
      \textbf{37.7}                                                                 & \textbf{63.0}                                                                        &
      \textbf{39.9}                                                                 & 48.8                                                                                 & 41.7                                                     & \textbf{45.2} &
      \textbf{43.0}                                                                 & \textbf{45.2}                                                                        & 31.7                                                     & \textbf{40.7} &
      \textbf{40.3}                                                                 & \textbf{28.8}                                                                        & \textbf{37.4}                                                              \\

      \bottomrule
    \end{tabular}
  }
  \label{tab:taskmap}
\end{table*}

\section{Experiments}

In line with previous work ~\cite{josephOpenWorldObject2021}~\cite{guptaOWDETROpenworldDetection2022}~\cite{wu2022uc}, we evaluate the performance of our framework using the Open World Evaluation Protocol ~\cite{josephOpenWorldObject2021} applied to the Pascal VOC and MS-COCO datasets. The combined datasets are divided into four tasks of 20 classes each. For each task, training images containing only the classes and bounding boxes for the 20 classes in the current task are used to train a detection model. The model is evaluated after each task using the Pascal VOC~\cite{everinghamPascalVisualObject2010} test and MS-COCO~\cite{linMicrosoftCOCOCommon2014} validation splits, with each class of the current and previous tasks treated as known, while any classes from future tasks are expected to be classified as unknown.

\noindent\textbf{Evaluation Metrics:} To measure the framework's detection performance on the known object classes in each task, we use the standard mean average precision metric (mAP). To measure its ability to detect unknown objects, we report the recall on the unknown class (U-Recall), and to measure its resilience to open set errors, we report the absolute open-set error (A-OSE) and Wilderness Impact (WI)~\cite{dhamijaOverlookedElephantObject2020}. Values are reported for all four tasks so that performance under incremental learning is apparent.

\noindent\textbf{Implementation:} Our implementation builds on the Faster-RCNN FPN implementation in the \texttt{detectron2}~\cite{wu2019detectron2} library and implements the new losses, heads, and algorithms as described in \cref{sec:ourarchitecture}. The metrics reporting code is adapted from ~\cite{guptaOWDETROpenworldDetection2022}. We use $0.69$ for the object score threshold $\theta_{obj}$, $0.5$ for the class score threshold $\theta_{cls}$, and $0.05$ for the low class confidence threshold $\theta_{conf}$. $300$ exemplars per class are used for incremental learning.
\begin{figure}[tpb]
  \centering
  \includegraphics[width=0.49\linewidth]{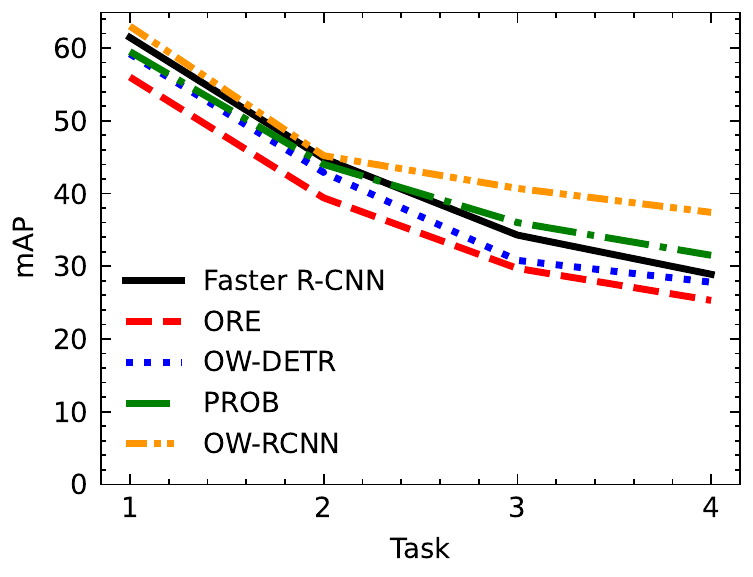}
  \includegraphics[width=0.49\linewidth]{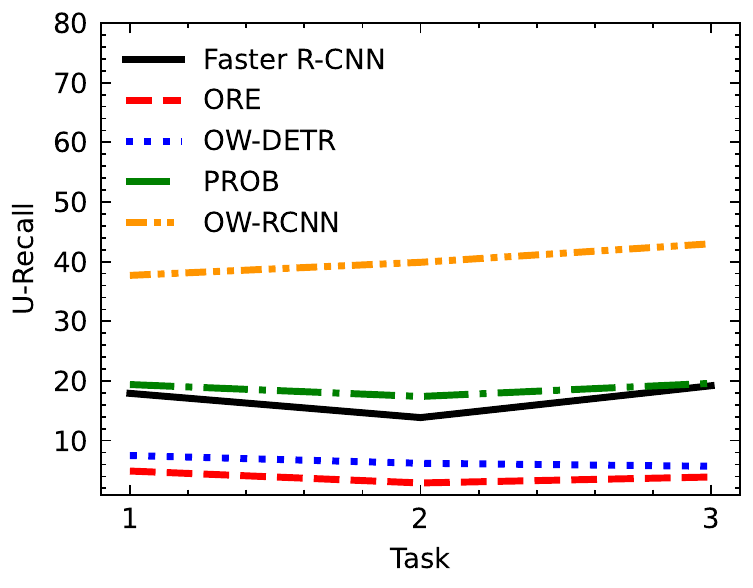}

  \caption{\textbf{Performance trends over the four tasks on MS-COCO.} Our framework shows a reduced rate of decrease in the mAP metric over time (left), while continuing to improve the ability to identify unknowns (U-Recall) (right).}
  \label{fig:trend}
\end{figure}
\subsection{Comparison with State-of-the-art}

We compare our performance with previously published OWOD methods. Following ~\cite{guptaOWDETROpenworldDetection2022}, the results reported for ORE exclude the energy-based unknown identification stage (EBUI), which requires unknown objects to have been annotated in the training data. For similar reasons, UC-OWOD is presented without its unknown clustering refinement (UCR) enhancement stage.

Motivated by Dhamija et al.'s~\cite{dhamijaOverlookedElephantObject2020a} observation that a non-open-world Faster R-CNN exhibited some robustness to open-set scenarios, we propose a simple Faster R-CNN-based baseline for our open-world detection task. Specifically, during inference, we treat all detections of known classes with a confidence score lower than 0.2 as unknown objects. The proposed baseline is trained using the same exemplar replay strategy as previous work, without any additional modifications to specifically tackle the challenges identified.

\begin{figure*}[!htb]
  \centering
  \includegraphics[align=b,width=0.24\linewidth]{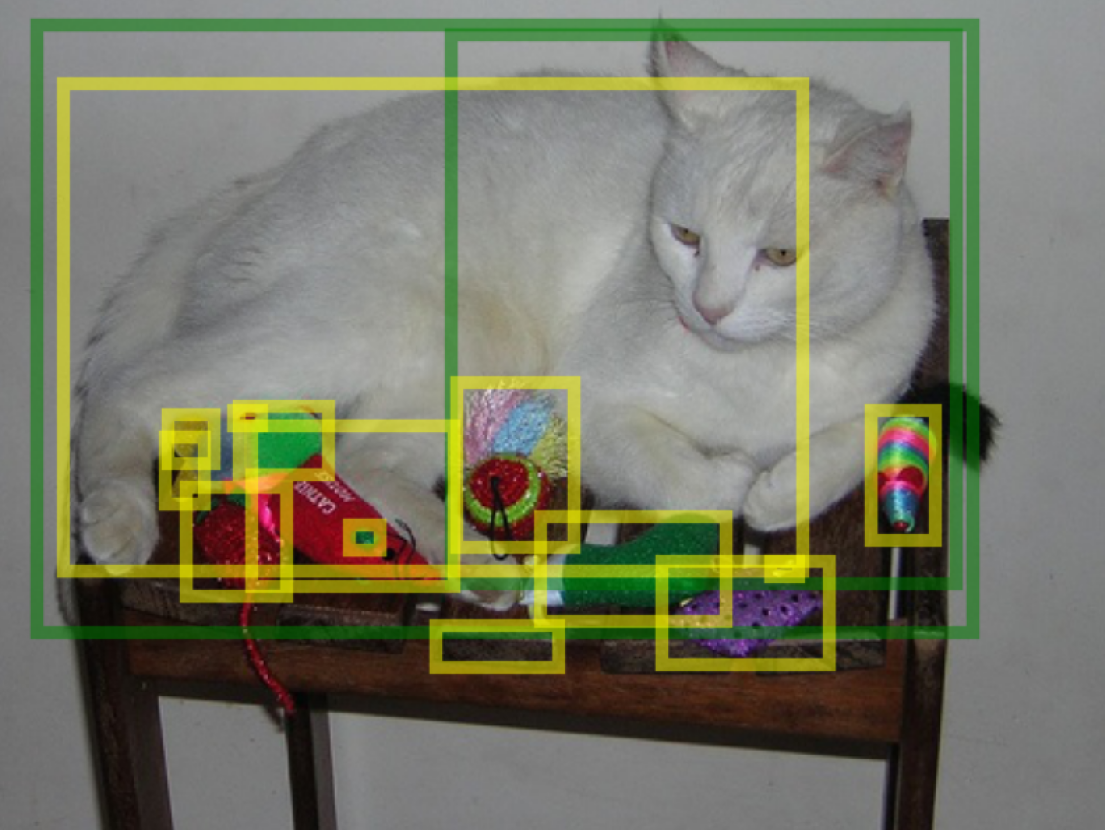}
  \includegraphics[align=b,width=0.24\linewidth]{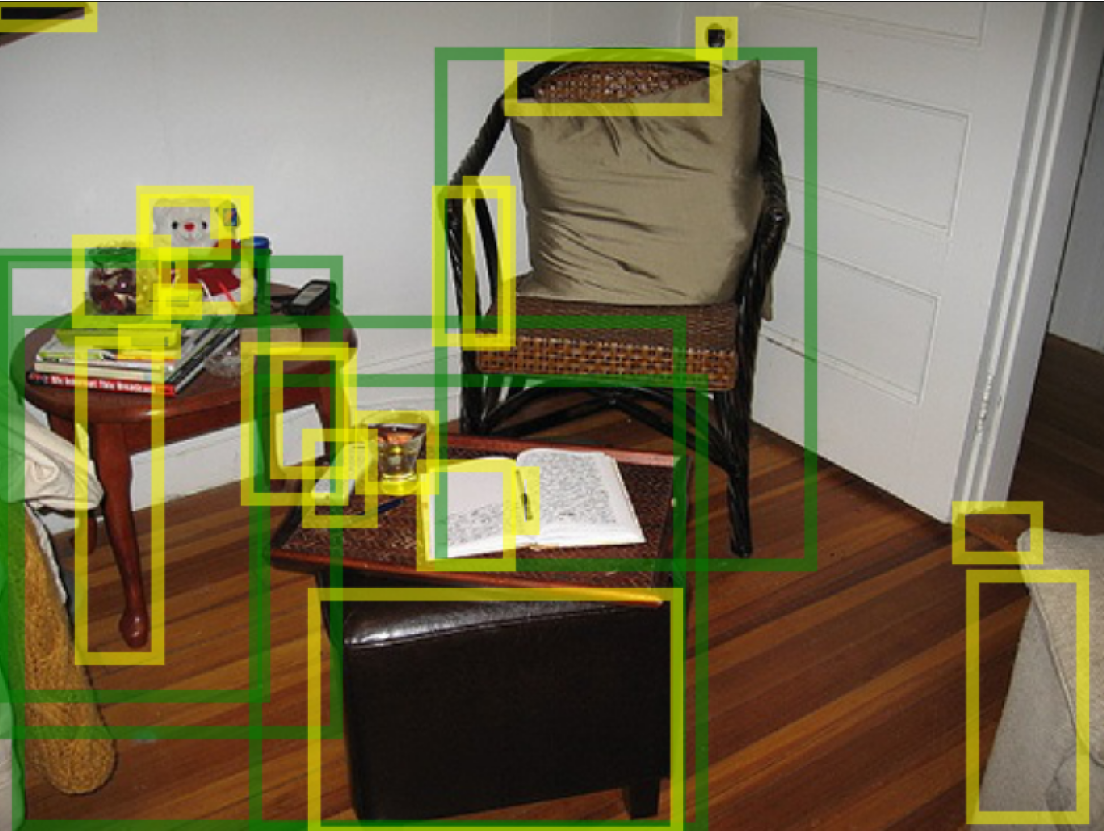}
  \includegraphics[align=b,width=0.24\linewidth]{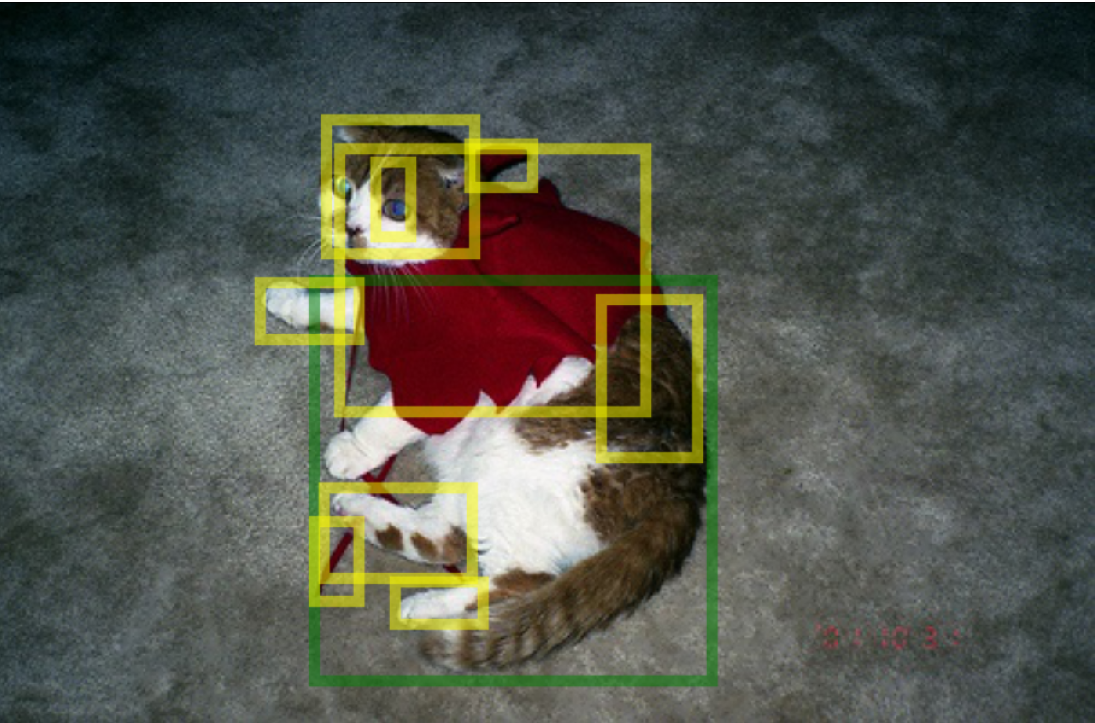}
  \includegraphics[align=b,width=0.24\linewidth]{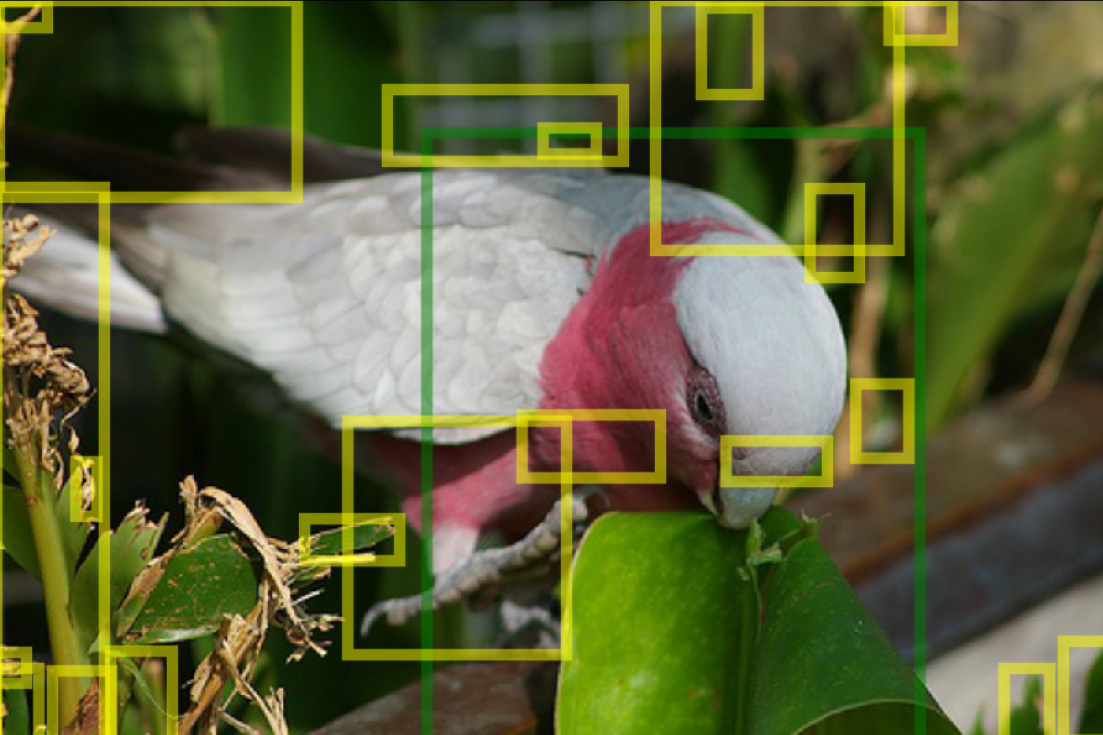}
  \caption{\textbf{Qualitative results on the MS-COCO test set.} The known class (\textcolor{green}{green}) and unknown class (\textcolor{yellow}{yellow}) detections from our framework have been overlaid. We highlight our detector's ability to isolate many unique objects in a scene (left, centre left). However, the detector reveals a couple of failure modes: it can tend to select parts of an object, such as legs or eyes (centre right), or sections of noisy background (right). }
  \label{fig:qualitative}
\end{figure*}

Our results are presented in \cref{tab:taskmap} and \cref{tab:taskaose}, with the results for other frameworks reported in their respective papers. We note that our naive baseline performs remarkably well compared to existing methods; however, it does not outperform \owrcnn{} which shows absolute performance gains of between 18.6 and 20.7\% on the U-Recall metric compared to the current state-of-the-art in U-Recall~\cite{maCATLoCalizationIdentificAtion2023}, highlighting its improved ability to detect unknown objects in a scene. We also observed absolute improvements in the mAP scores between 1.2 and 5.9\% compared to previous methods. The U-Recall shows increased performance in the three tasks (\cref{fig:trend}), reflecting an improved ability to detect unknown objects as the number of known classes increases. Beyond absolute gains in mAP, we observe an improved balance between previously known and current known class mAP when compared against the other frameworks. Qualitative results of these improvements are presented in \cref{fig:qualitative}.

Our framework shows an improvement in the absolute open-set error metric (A-OSE), reducing it by up to 52\% and improving wilderness impact (WI) in most tasks.

A side effect of our change to the continual learning strategy is the reduced number of images required (\cref{tab:taskaose}) to achieve better detection results. This reduction in training images also results in a shorter training time.

The improvements across all metrics, including significant improvements in U-Recall and A-OSE, highlight the impact of our proposed contributions and the combined effect of methodically addressing each challenge presented by the open-world object detection task.

\begin{table*}[pthp]\setlength{\tabcolsep}{6pt}
  \centering
  \caption{\textbf{Performance comparison on MS-COCO for OWOD.} Our framework shows a major improvement in the ability to avoid open set errors compared to previously published methods, along with competitive results on the Wilderness Impact (WI) metric. The proposed training regime allows our method to reduce the amount of data on which the detection network is trained.}\vspace{0.2cm}
  \resizebox{0.9\textwidth}{!} {
    \begin{tabular}{@{}lcccccccccc@{}}
      \toprule

      Task IDs ($\rightarrow$)                                                   &
      \multicolumn{3}{c}{Task 1}                                                 &
      \multicolumn{3}{c}{Task 2}                                                 &
      \multicolumn{3}{c}{Task 3}                                                 &
      Task 4                                                                                                                           \\

      \cmidrule(lr){2-4}\cmidrule(lr){5-7}\cmidrule(lr){8-10}\cmidrule(l){11-11} &
      A-OSE                                                                      & WI              & Images         &
      A-OSE                                                                      & WI              & Images         &
      A-OSE                                                                      & WI              & Images         &
      Images                                                                                                                           \\

                                                                                 & ($\downarrow$)  & ($\downarrow$) & ($\downarrow$) &
      ($\downarrow$)                                                             & ($\downarrow$)  & ($\downarrow$) &
      ($\downarrow$)                                                             & ($\downarrow$)  & ($\downarrow$) &
      ($\downarrow$)                                                                                                                   \\

      \midrule

      Faster R-CNN\scriptsize{(+threshold)}                                      &
      \textbf{3925}                                                              & 0.0503          & 16551          &
      \textbf{1515}                                                              & 0.0250          & 46606          &
      \textbf{990}                                                               & 0.0164          & 40901          &
      42253                                                                                                                            \\

      \midrule

      ORE - EBUI~\cite{josephOpenWorldObject2021}                                &
      10459                                                                      & 0.0621          & 16551          &
      10445                                                                      & 0.0282          & 46606          &
      7990                                                                       & 0.0211          & 40901          &
      42253                                                                                                                            \\

      OW-DETR~\cite{guptaOWDETROpenworldDetection2022}                           &
      10240                                                                      & 0.0571          & 16551          &
      8441                                                                       & 0.0278          & 46606          &
      6803                                                                       & 0.0156          & 40901          &
      42253                                                                                                                            \\

      UC-OWOD~\cite{wu2022uc}                                                    &
      9294                                                                       & -               & 16551          &
      5062                                                                       & -               & 46606          &
      3801                                                                       & -               & 40901          &
      42253                                                                                                                            \\

      2B-OCD~\cite{wuTwobranchObjectnesscentricOpen2022}                         &
      -                                                                          & \textbf{0.0480} & 16551          &
      -                                                                          & \textbf{0.0160} & 46606          &
      -                                                                          & \textbf{0.0137} & 40901          &
      42253                                                                                                                            \\

      CAT~\cite{maCATLoCalizationIdentificAtion2023}                             &
      7070                                                                       & 0.0581          & 16551          &
      5000                                                                       & 0.0275          & 46606          &
      4241                                                                       & 0.0173          & 40901          &
      42253                                                                                                                            \\

      PROB~\cite{zoharPROBProbabilisticObjectness2022}                           &
      5195                                                                       & 0.0569          & 16551          &
      6452                                                                       & 0.0344          & 46606          &
      4641                                                                       & 0.0151          & 40901          &
      42253                                                                                                                            \\

      \midrule

      \textbf{\owrcnn{} (Ours)}                                                  &
      6957                                                                       & 0.0524          & 16551          &
      2487                                                                       & 0.0233          & \textbf{6333}  &
      1820                                                                       & 0.0165          & \textbf{8746}  &
      \textbf{10966}                                                                                                                   \\

      \bottomrule
    \end{tabular}
  }
  \label{tab:taskaose}
\end{table*}

\begin{table*}[!bhtp]\setlength{\tabcolsep}{6pt}
  \caption{\textbf{The impact of each of our components.} The comparison shows the changes in known class detection performance (mAP), absolute open-set errors (A-OSE), and the ability to detect unknown objects (U-Recall). The harmonic mean of the mAP of the previously known and current known classes (\F1i score~\cite{pengSIDIncrementalLearning2020}) is included to monitor the impact of incremental learning on the balance of detection performance between previous and current classes.}\vspace{0.2cm}
  \resizebox{\textwidth}{!} {
    \begin{tabular}{@{}lcccccccccc@{}}
      \toprule

      Task IDs ($\rightarrow$)                                                    &
      \multicolumn{3}{c}{Task 1}                                                  &
      \multicolumn{3}{c}{Task 2}                                                  &
      \multicolumn{3}{c}{Task 3}                                                  &
      Task 4
      \\

      \cmidrule(lr){2-4}\cmidrule(lr){5-7}\cmidrule(lr){8-10}\cmidrule(lr){11-11} &
      U-Recall                                                                    & A-OSE          & mAP          &
      U-Recall                                                                    & A-OSE          & mAP/\F1i     &
      U-Recall                                                                    & A-OSE          & mAP/\F1i     &
      mAP/\F1i                                                                                                      \\

                                                                                  &
      ($\uparrow$)                                                                & ($\downarrow$) & ($\uparrow$) &
      ($\uparrow$)                                                                & ($\downarrow$) & ($\uparrow$) &
      ($\uparrow$)                                                                & ($\downarrow$) & ($\uparrow$) &
      ($\uparrow$)                                                                                                  \\

      \midrule
      Faster R-CNN + Fine-tune                                                    &
      0.0                                                                         & 8955           & 62.9         &
      0.0                                                                         & 4323           & 46.6 / 46.6  &
      0.0                                                                         & 3572           & 36.5 / 36.8  &
      31.0 / 30.3                                                                                                   \\

      + Fine-tune only                                                            &
      0.0                                                                         & 6510           & 64.9         &
      0.0                                                                         & 2518           & 47.6 / 47.5  &
      0.0                                                                         & 1634           & 42.4 / 38.9  &
      39.4 / 35.2                                                                                                   \\

      + Class agnostic detection                                                  &
      37.6                                                                        & 7061           & 62.9         &
      38.9                                                                        & 3011           & 45.5 / 45.2  &
      43.3                                                                        & 2202           & 40.5 / 37.2  &
      37.3 / 33.2                                                                                                   \\

      + Probability-based unknown detection                                       &
      37.7                                                                        & 6957           & 63.0         &
      39.4                                                                        & 2573           & 45.4 / 45.2  &
      44.0                                                                        & 1974           & 40.3 / 37.0  &
      37.1 / 33.1                                                                                                   \\

      + Prior task class handling (\owrcnn{})                                     &
      \multicolumn{3}{c}{\textit{Not used in task 1}}                             &
      39.9                                                                        & 2487           & 45.2 / 44.9  &
      43.0                                                                        & 1820           & 40.7 / 37.2  &
      37.4 / 33.6                                                                                                   \\
      \bottomrule
    \end{tabular}
  }
  \label{tab:ablation}
\end{table*}

\section{Ablation study}
\label{sec:ablation}

To show the contribution of each of our proposed changes to overall performance, we report the results of progressively adding each component. The results are presented in \cref{tab:ablation}. They show an improvement in detection accuracy as the new training regime is added, even in the first task where we can now train the network for more epochs without impacting future learning by over-specialising. A trade-off is revealed between the ability to detect unknowns and a drop in mAP and A-OSE performance when we introduce and use the new localisation-quality heads for unknown object detection and class-agnostic region proposals. Probability-based unknown detection reduces the absolute open-set error while also slightly improving the U-Recall metric on most tasks. The handling of prior task classes has only a minor impact on task $2$, but results in an improved mAP for tasks $3$ and $4$.

\section{Conclusion}
By examining open-world object detection through the lens of its three core challenges, we have proposed a new open-world object detection framework. It achieves a new state-of-the-art for the open-world evaluation protocol using MS-COCO, with performance gains in object detection, open-set error reduction, unknown object recall, and training time. It represents a holistic approach to open-world object detection and provides a strong baseline for future research.

\clearpage
{\small
  \bibliographystyle{ieee_fullname}
  \bibliography{ref-small}
}

\clearpage

\appendix

\section{Further Evaluation}

\subsection{Alternative MS-COCO Task Split}

Gupta \etal~\cite{guptaOWDETROpenworldDetection2022} propose an alternative task split for the MS-COCO dataset under the open-world evaluation protocol. Their proposal organises the tasks by non-overlapping super-categories (see \cref{tab:hardsplit}) mitigating possible data leakage between tasks. It uses only images from the MS-COCO dataset for training and testing, removing the dependency on the Pascal VOC dataset.

\begin{table}[!htp]
  \centering
  \resizebox{\linewidth}{!} {
    \begin{tabular}{llc}
      \toprule
      Task & Super-categories                            & Training Images \\
      \midrule
      1    & Animals, Person, Vehicles                   & 89490           \\
      2    & Appliances, Accessories, Outdoor, Furniture & 55870           \\
      3    & Food, Sport                                 & 39402           \\
      4    & Electronic, Indoor, Kitchen                 & 38903           \\
      \bottomrule
    \end{tabular}}
  \caption{The proposed MS-COCO task split}

  \label{tab:hardsplit}
\end{table}

We compare the performance of \owrcnn{} with OW-DETR and ORE under this new split and present the results in \cref{tab:newsplitperf}. \owrcnn{} shows an similar increase in U-Recall as in the original task split. The mAP performance of our method is improved over the previous Faster R-CNN-based ORE. Our method also outperforms the transformer-based OW-DETR in the final two tasks, however OW-DETR achieves a greater mAP in the first two tasks.

\begin{table*}[!thpb]\setlength{\tabcolsep}{2pt}
  \caption{\textbf{State-of-the-art for the Open World Evaluation Protocol on the alternate MS-COCO split.} We compare our framework with other implementations on the alternative task class split as proposed by ~\cite{guptaOWDETROpenworldDetection2022}. Following ~\cite{guptaOWDETROpenworldDetection2022}, we report the performance of ORE~\cite{josephOpenWorldObject2021} without the energy-based unknown identifiers. The mAP figures are presented for previously known classes, those that were learnt as part of the current task, and for all known classes. The unknown class recall (U-Recall) represents the portion of unknown objects retrieved by the model. \owrcnn{} shows large improvements in U-Recall however overall detection performance is lower than CAT and PROB for this regime which has 5x the training data for task 1. However this gap closes in the later tasks.
  }\vspace{0.2cm}
  \resizebox{\textwidth}{!} {
    \begin{tabular}{@{}lcccccccccccccccc@{}}
      \toprule

      Task IDs ($\rightarrow$)                                                      &
      \multicolumn{2}{c}{Task 1}                                                    &
      \multicolumn{4}{c}{Task 2}                                                    &
      \multicolumn{4}{c}{Task 3}                                                    &
      \multicolumn{3}{c}{Task 4}                                                                                                                                                                                                                        \\

      \cmidrule(lr){2-3}\cmidrule(lr){4-7}\cmidrule(lr){8-11}\cmidrule(lr){12-14}   &
      U-Recall                                                                      & \multicolumn{1}{c}{mAP ($\uparrow$)}                                                 &
      U-Recall                                                                      & \multicolumn{3}{c}{mAP ($\uparrow$)}                                                 &
      U-Recall                                                                      & \multicolumn{3}{c}{mAP ($\uparrow$)}                                                 &
      \multicolumn{3}{c}{mAP ($\uparrow$)}                                                                                                                                                                                                              \\

      \cmidrule(lr){3-3} \cmidrule(lr){5-7} \cmidrule(lr){9-11}\cmidrule(lr){12-14} &
      ($\uparrow$)                                                                  & \begin{tabular}[c]{@{}c}Current \\ known\end{tabular} &
      ($\uparrow$)                                                                  & \begin{tabular}[c]{@{}c@{}}Previously\\  known\end{tabular}                          & \begin{tabular}[c]{@{}c@{}}Current \\ known\end{tabular} & Both          &
      ($\uparrow$)                                                                  & \begin{tabular}[c]{@{}c@{}}Previously \\ known\end{tabular}                          & \begin{tabular}[c]{@{}c@{}}Current \\ known\end{tabular} & Both          &
      \begin{tabular}[c]{@{}c@{}}Previously \\ known\end{tabular}                   & \begin{tabular}[c]{@{}c@{}}Current \\ known\end{tabular}                             & Both                                                                       \\

      \midrule
      ORE - EBUI~\cite{josephOpenWorldObject2021}                                   &
      1.5                                                                           & 61.4                                                                                 &
      3.9                                                                           & 56.5                                                                                 & 26.1                                                     & 40.6          &
      3.6                                                                           & 38.7                                                                                 & 23.7                                                     & 33.7          &
      33.6                                                                          & 26.3                                                                                 & 31.8                                                                       \\

      OW-DETR~\cite{guptaOWDETROpenworldDetection2022}                              &
      5.7                                                                           & 71.5                                                                                 &
      6.2                                                                           & 62.8                                                                                 & 27.5                                                     & 43.8          &
      6.9                                                                           & 45.2                                                                                 & 24.9                                                     & 38.5          &
      38.2                                                                          & 28.1                                                                                 & 33.1                                                                       \\

      CAT~\cite{maCATLoCalizationIdentificAtion2023}                                &
      \textbf{24.0}                                                                 & \textbf{74.2}                                                                        &
      23.0                                                                          & \textbf{67.6}                                                                        & 35.5                                                     & \textbf{50.7} &
      24.6                                                                          & \textbf{51.2}                                                                        & \textbf{32.6}                                            & \textbf{45.0} &
      \textbf{45.4}                                                                 & \textbf{35.1}                                                                        & \textbf{42.8}                                                              \\

      PROB\cite{zoharPROBProbabilisticObjectness2022}                               &
      17.6                                                                          & 73.4                                                                                 &
      22.3                                                                          & 66.3                                                                                 & 36.0                                                     & 50.4          &
      24.8                                                                          & 47.8                                                                                 & 30.4                                                     & 42.0          &
      42.6                                                                          & 31.7                                                                                 & 39.9                                                                       \\

      \midrule

      \textbf{\owrcnn{} (Ours)}                                                     &
      23.9                                                                          & 68.9                                                                                 &
      \textbf{33.3}                                                                 & 49.6                                                                                 & \textbf{36.7}                                            & 41.9          &
      \textbf{40.8}                                                                 & 42.3                                                                                 & 30.8                                                     & 38.5          &
      39.4                                                                          & 32.2                                                                                 & 37.7                                                                       \\

      \bottomrule
    \end{tabular}
  }
  \label{tab:newsplitperf}
\end{table*}

\subsection{Practical Mode Protocol}

A compelling aspect of open-world object detectors is that they are capable of collecting their own future training data when identifying currently unknown objects. Although it is not possible for them to provide the correct class labels for such unknown objects, they can locate them within a scene and provide a score ranking. In an ideal application, an oracle would only be asked to provide the class labels for the unknown regions detected and localised by the model during its operation. This mode of operation results in a practical reduction in the labelling workload. To measure an open-world object detector's suitability to this workflow, we propose a new "Practical Mode" addition to the open-world evaluation protocol. Unlike the original evaluation protocol, which provides each task with a set of training data that is independent of the performance of the previous task, our proposed protocol uses the model of the previous task $\mathcal{M}^{t-1}$ to find the regions containing unknown objects from the training data of the current task. To model classification by an oracle, each detected unknown region that overlaps a ground truth region is annotated with that region's class. The unknown regions that were assigned a class are then used as the only training data $D^{t}$ for the new classes of the current task. This process is repeated for the remaining tasks.

\textbf{Dataset:} As a concrete implementation of this protocol, we use the alternative MS-COCO split proposed by ~\cite{josephOpenWorldObject2021}, which organises the tasks by non-overlapping super-categories (\cref{tab:hardsplit}) mitigating possible data leakage between tasks.

\textbf{Metrics:} We report a set of metrics reflecting the goals of the open world object detection task: mean average precision (mAP) to measure its ability to detect known classes, recall on the unknown class (U-Recall) to measure its ability to detect unknown objects, absolute open set error (A-OSE) to measure confusion introduced by open set, and the harmonic mean of the mAP of the previously known and current known classes (\F1i score~\cite{pengSIDIncrementalLearning2020}) to monitor the impact of incremental learning.

\textbf{Results:} We compare \owrcnn{} with the recent OW-DETR in \cref{tab:practicalmode} using the same metrics as the original open world evaluation protocol, with the average precision calculated using the Pascal VOC 2010 method. Unlike previous results reported in the paper, the OW-DETR results for this benchmark are obtained by running the source code provided by Gupta \etal~\cite{guptaOWDETROpenworldDetection2022}.

\begin{figure}
  \centering

  \includegraphics[width=1\linewidth]{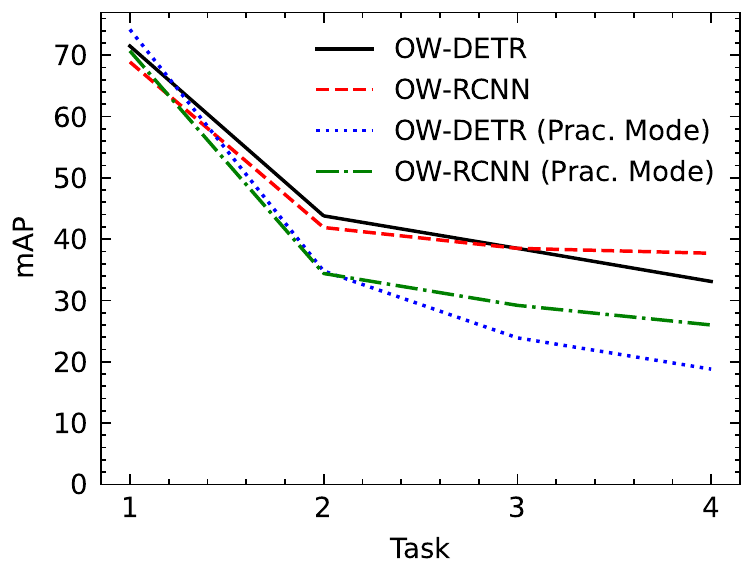}

  \caption{The impact of the introduction of "Practical Mode" on the mAP performance of \owrcnn{} and OW-DETR. \owrcnn{}'s improved U-Recall allows it to retain a higher mAP in later tasks.}
  \label{fig:pracmodeperf}
\end{figure}

\begin{table*}[!htpb]\setlength{\tabcolsep}{6pt}
  \caption{\textbf{Practical-Mode on the COCO benchmark.} We compare our framework with OW-DETR on our proposed "Practical Mode" addition to the evaluation protocol. \owrcnn{} continues to show improved unknown recall (U-Recall) and absolute open-set error (A-OSE). Although the mean average precision (mAP) scores are lower in the first two tasks, the \F1i score shows that \owrcnn{} produces a balanced performance between the prior and current tasks. By tasks $3$ and $4$, \owrcnn{} outperforms OW-DETR on all metrics, highlighting the impact of improved U-Recall when operating in this mode. }\vspace{0.2cm}
  \resizebox{\textwidth}{!} {
    \begin{tabular}{@{}lcccccccccc@{}}
      \toprule

      Task IDs ($\rightarrow$)                                                    &
      \multicolumn{3}{c}{Task 1}                                                  &
      \multicolumn{3}{c}{Task 2}                                                  &
      \multicolumn{3}{c}{Task 3}                                                  &
      \multicolumn{1}{c}{Task 4}                                                                                          \\

      \cmidrule(lr){2-4}\cmidrule(lr){5-7}\cmidrule(lr){8-10}\cmidrule(lr){11-11} &
      U-Recall                                                                    & A-OSE          & mAP                &
      U-Recall                                                                    & A-OSE          & mAP/\F1i           &
      U-Recall                                                                    & A-OSE          & mAP/\F1i           &
      mAP/\F1i                                                                                                            \\

                                                                                  &
      ($\uparrow$)                                                                & ($\downarrow$) & ($\uparrow$)       &
      ($\uparrow$)                                                                & ($\downarrow$) & ($\uparrow$)       &
      ($\uparrow$)                                                                & ($\downarrow$) & ($\uparrow$)       &
      ($\uparrow$)                                                                                                        \\

      \midrule

      OW-DETR                                                                     &
      4.4                                                                         & 13257          & \textbf{74.2}      &
      3.4                                                                         & 17624          & \textbf{34.8}/15.1 &
      5.8                                                                         & 9160           & 23.9/11.3          &
      18.8/10.3
      \\
      \midrule

      \textbf{\owrcnn{} (Ours)}                                                   &
      \textbf{23.9}                                                               & \textbf{2726}  & 70.7               &
      \textbf{38.8}                                                               & \textbf{1873}  & 34.4/\textbf{34.4} &
      \textbf{39.3}                                                               & \textbf{1127}  & \textbf{29.2/28.6} &
      \textbf{26.0/26.0}                                                                                                  \\
      \bottomrule
    \end{tabular}
  }

  \label{tab:practicalmode}
\end{table*}

\subsection{Unknown Precision}

Relying on U-Recall to measure the ability of the detector to locate unknowns provides the potential for a detector to show increased performance simply by proposing large numbers of regions as unknown, increasing the chance of locating an unknown object that is labelled in the test set. Previous work ~\cite{josephOpenWorldObject2021, guptaOWDETROpenworldDetection2022} points out that since unknown objects are not exhaustively labelled in the test dataset, it is not possible to get an accurate unknown precision metric which would identify this behaviour. However, while not all unknowns in the test dataset are labelled, the known unknowns (future task classes) are labelled and are already used to compute the U-Recall metric. We can examine the precision of the detectors on these known unknown classes and while it would be unlikely to achieve $100\%$ due to detections of unknown objects outside the MS-COCO class set, it should provide an intuition as to whether any U-Recall is due to large numbers of false positive detections. \cref{tab:uprecision} shows the performance of our execution of OW-DETR and \owrcnn{} on the first task of the alternative MS-COCO split.

\begin{table}[!htpb]\small
  \centering
  \caption{\textbf{Recall performance on Task 1 of alternative MS-COCO split.} We show precision and recall performance on the unknown items from the MS-COCO test set. Precision scores will be much lower than $100$ since there are many detections for unknown objects outside of those labelled in the test set, however both frameworks face a similar problem and their relative scores should give an intuition as to the precision of their detectors. \owrcnn{} shows a large increase in recall without a corresponding drop in precision.}\vspace{0.2cm}
  \begin{tabular}{lccc}
    \toprule
    Framework                 & Recall         & Precision     & $\text{AP}_{50}$ \\
    \midrule
    OW-DETR                   & 4.38           & 0.49          & 0.02             \\
    \midrule
    \textbf{\owrcnn{} (ours)} & \textbf{23.88} & \textbf{2.25} & \textbf{1.10}    \\
    \bottomrule
  \end{tabular}

  \label{tab:uprecision}
\end{table}

\section{Threshold Ablation}
\subsection{Objectness Threshold ($\theta_{obj}$)}

\begin{figure}
  \centering

  \includegraphics[width=0.49\linewidth]{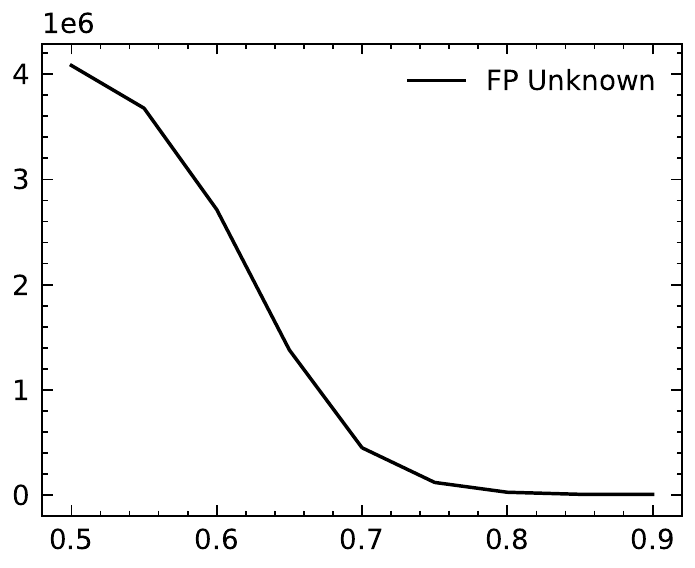}
  \includegraphics[width=0.49\linewidth]{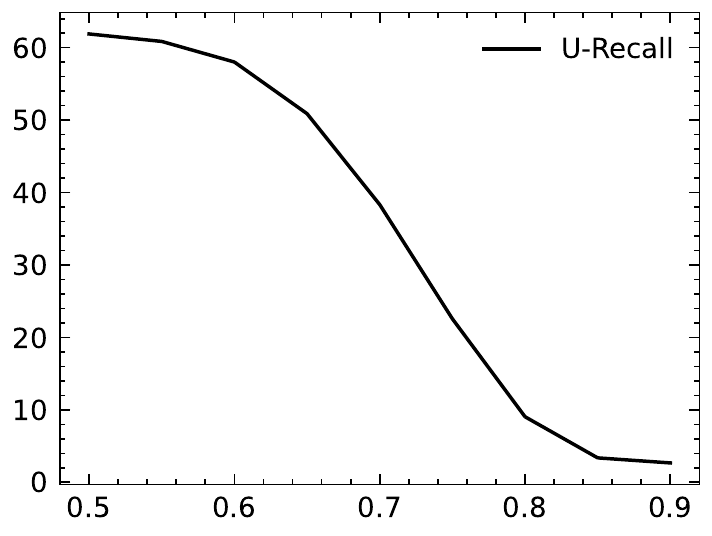}

  \caption{\textbf{Objectness Threshold Ablation.} We show a drop in false positive unknown detections with the corresponding drop in recall as we increase the threshold. Thresholds around $0.65$ to $0.75$ retain high unknown recall while reducing false positive detections.}
  \label{fig:ablateobj}
\end{figure}

The objectness threshold specifies the objectness score above which proposals are considered objects. Since the objectness scores are trained only on ground truth objects, the majority of the training input has a non-zero objectness score. A suitable value for the objectness threshold $\theta_{obj}$ is one that provides a good trade-off between the number of false positive detections of unknown objects and its recall on the unknown objects. We tested multiple threshold values in task $2$ of the MS-COCO for OWOD protocol and present the results in \cref{fig:ablateobj}. We find that low values of the threshold result in large numbers of false positive detections. This number decreases as the threshold increases, with the rate of decrease slowing just before $0.7$. Unknown recall also shows a corresponding decrease; however, it remains relatively high at the $0.7$ value, slowing its rate of decrease at about $0.8$. Thresholds between $0.65$ and $0.75$ provide a balanced performance between false positives and recall. We use $0.69$ as the threshold in our implementation so that regions with objectness scores of $0.7$ and above are considered objects.

\subsection{Confidence Threshold ($\theta_{conf}$)}

\begin{figure}
  \centering

  \includegraphics[width=\linewidth]{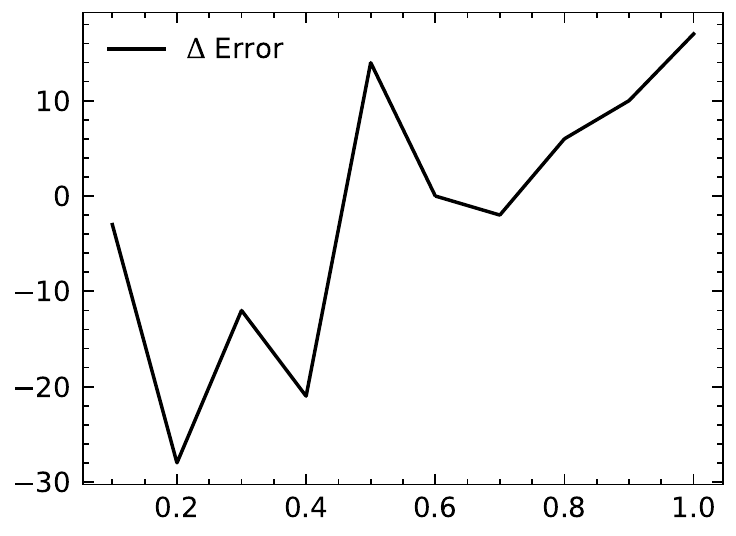}

  \caption{\textbf{Confidence Threshold Ablation.} The difference ($\Delta$ Error) in the reduction of absolute open-set errors reduced and the increase in false positive detections of a known object as unknown, observed for each increment of the threshold. After $0.5$ subsequent increases in the confident threshold start to introduce more errors than it removes. }
  \label{fig:ablateconf}
\end{figure}

The confidence threshold $\theta_{conf}$ provides a mechanism to avoid converting high scoring detections of known classes into the unknown class when reducing open-set errors using the Gaussian mixture models $G_k$. We examine the relationship between the confidence threshold and the absolute open-set error (A-OSE) compared to the number of known objects that were erroneously detected as unknown in \cref{fig:ablateconf}. Although the overall number of errors is reduced for threshold values below $0.5$, for higher values there is an increase in the errors introduced by misidentifying known objects as unknown that exceeds the amount of absolute open set errors reduced by going to that threshold value.

\section{Limitations}

While providing performance gains across most metrics, we have identified some limitations in our framework.
\begin{itemize}
  \item By only using the exemplar set to train the detection model for each task, critical training images may be missed. However, a more intelligent sampling regime may reduce the risk of this occurring.
  \item The Gaussian mixture models used for open-set error detection need to be trained on a minimum number of samples to ensure their accuracy. Their performance may be affected in low-data scenarios.
  \item The class scores returned for unknown objects are of a different scale from the score for the known classes and cannot be interpreted as probabilities. Implementations of the framework will need to use a different confidence threshold for unknown objects compared to the one used for known classes.
\end{itemize}

\end{document}